\documentclass{article}
\usepackage{arxiv}

\usepackage{graphicx} 
\usepackage{amsmath} 
\usepackage{amssymb}  
\usepackage{caption}
\usepackage{subcaption}
\usepackage{hyperref}
\newcommand{\eg}{\emph{e.g.}, }
\newcommand{\ie}{\emph{i.e.}, } 
\newcommand{\etal}{\emph{et al.} } 

\title{\LARGE \bf
Visual Diver Face Recognition \\for Underwater Human-Robot Interaction
}

\author{Jungseok Hong$^{1}$, Sadman Sakib Enan$^{2}$, Christopher Morse$^{3}$ and Junaed Sattar$^{4}$
\thanks{The authors are with the Department of Computer Science and Engineering and the Minnesota Robotics Institute,
        University of Minnesota Twin Cities, Minneapolis, MN, USA.
        {\tt\small \{$^{1}$jungseok,$^{2}$enan0001,$^{3}$morse164,$^{4}$junaed\}@umn.edu}}%
\thanks{*This work was supported by the US National Science Foundation Award IIS-\#1845364, the UMII-MnDRIVE Fellowship, and the MnRI Seed Grant. The first two authors made equal contribution to the work and should both be cited as first author (ex. \textit{Hong and Enan et al.})}%
}

\begin{document}

\maketitle
\thispagestyle{empty}
\pagestyle{empty}

\begin{abstract}

This paper presents a deep-learned facial recognition method for underwater robots to identify scuba divers. Specifically, the proposed method is able to recognize divers underwater with faces heavily obscured by scuba masks and breathing apparatus. Our contribution in this research is towards robust facial identification of individuals under significant occlusion of facial features and image degradation from underwater optical distortions. With the ability to correctly recognize divers, autonomous underwater vehicles (AUV) will be able to engage in collaborative tasks with the correct person in human-robot teams and ensure that instructions are accepted from only those authorized to command the robots. We demonstrate that our proposed framework is able to learn discriminative features from real-world diver faces through different data augmentation and generation techniques. Experimental evaluations show that this framework achieves a 3-fold increase in prediction accuracy compared to the state-of-the-art (SOTA) algorithms and is well-suited for embedded inference on robotic platforms.

\end{abstract}

\section{Introduction} \label{intro}

Autonomous underwater robots have seen increasing use in recent years in a plethora of applications, primarily driven by advances in cognitive abilities and computational resources (\eg navigation~\cite{xanthidis2020navigation,manderson2020vision}, detection~\cite{fulton2019robotic}). However, the underwater domain poses significant challenges to perception, locomotion, communication, and interaction. These include attenuation of electromagnetic signals, optical distortions, and control in six-degrees-of-freedom environments, leading to a significant reduction of situational awareness. Underwater human-robot teams have also seen some exciting advances of late. Generally, human-robot teamwork enables robots to rely on human cognition and instructions~\cite{islam2018dynamic}; it also allows humans to exploit robot autonomy for more efficient task completion and improve overall safety in human-robot interaction (HRI), owing to improved situational awareness. Yet, underwater human-robot communication remains an open challenge, particularly when robots are required to accept instructions from only a few select individuals. Robots need to \textit{uniquely identify divers} under similar wearables, facial occlusions, and degraded underwater optics, which are significant issues. Misidentifying divers may cause mission failure, pose substantial risks, and also compromise secure human-robotic interaction. The latter concern is significant in any underwater mission (\eg pipeline and cable inspections, search and rescue) but is particularly crucial to maritime defense applications. Previous work by Xia \etal\cite{xia2019visual} and de Langis \etal\cite{de2020realtime} have made some inroads in uniquely identifying underwater scuba divers from their swimming gaits. Still, these algorithms are inapplicable for close-proximity interactions as they require the entire body to be visible and in motion.

Vision underwater has been used extensively (\eg\cite{fulton2019robotic, islam2019understanding, weidner2017underwater}), despite being subjected to optical distortions. Particularly in the case of underwater HRI, there is sufficient light (natural or otherwise) available, as divers operate at littoral, shallower regions usually not exceeding approximately $25-30$ \textit{meters} of depth. Moreover, recent work in underwater image enhancement~\cite{islam2020sesr} have made it possible to use vision even more robustly; however, they are still faced with challenges in facial recognition tasks, as most features required by the SOTA algorithms~\cite{deng2019arcface, wang2018cosface} are obscured for divers by their breathing apparatus. With masks and regulators on, distinctive components of faces are obscured (\eg eyes, nose, mouths), which results in inferior quality feature representations. This issue is further intensified by the optical distortions present underwater~\cite{UnderwaterOptics}. As a result, the SOTA algorithms fail to identify subjects reliably underwater.

\begin{figure}[t]
    \centering
    \includegraphics[width=0.99\linewidth]{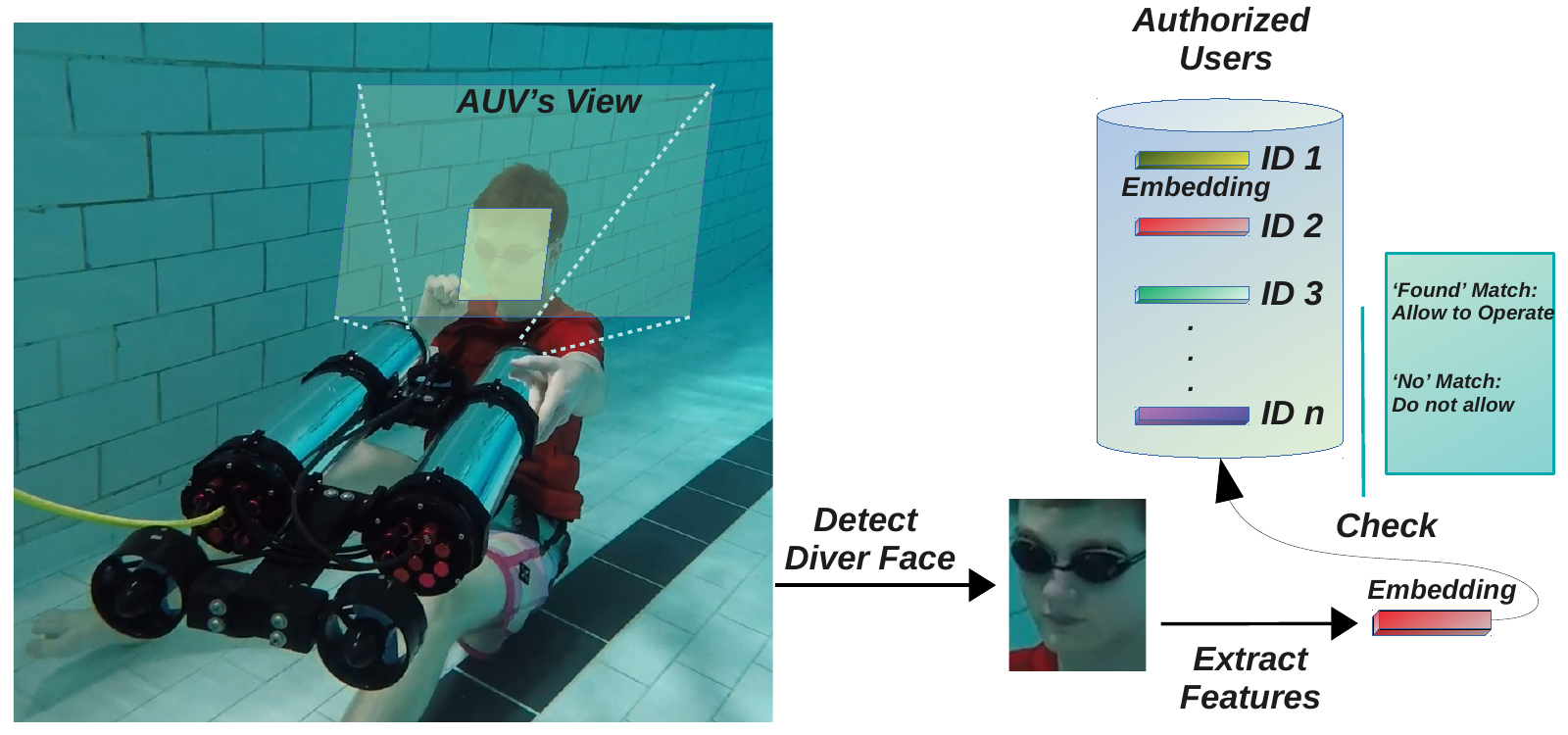}
    \caption{Demonstration of a diver face identification system for underwater human-robot interaction. The framework detects diver faces underwater and extracts discriminative feature embeddings which are matched against pre-computed embeddings of authorized users stored in a database.}
    \label{fig:diver_auth_system}
    \vspace{0mm}
\end{figure}

This work attempts to address this rather daunting challenge of underwater facial recognition of divers, as demonstrated in Fig.~\ref{fig:diver_auth_system}. Specifically, we propose an approach for robust face recognition when divers' faces are mostly obscured by masks and regulators (or snorkels and other breathing apparatus). To the best of our knowledge, our work is the first to address this unique challenge of diver recognition through facial features for underwater HRI tasks.

This paper makes the following contributions:
\begin{enumerate}
\item we develop a data augmentation technique to create realistic diver faces from non-diver faces and construct a dataset that includes both diver and non-diver faces.
\item also, we present a diver face identification system that can recognize scuba divers.
\item furthermore, we perform several quantitative and qualitative experiments that validate the proposed system's effectiveness. Finally, we analyze the practical feasibility of this framework for mobile platforms.
\end{enumerate}
\section{Related Work} \label{rel_work}

Recent advances (\eg hand gestures~\cite{islam2019understanding}, fiducial tags~\cite{dudek2007visual}) in underwater HRI have allowed divers to communicate with robots without using a tethered underwater control device~\cite{verzijlenberg2010swimming}. However, the problem of identifying a diver during the communication has not been well studied. With the advances in vision-based communication between underwater robots and divers~\cite{islam2019understanding}, underwater human-robot joint missions (\ie multi human-robot interaction or M/HRI) have become feasible. Thus, the ability for robots to identify individual divers in such teams have become imperative. Existing vision-based diver identification work~\cite{xia2019visual,de2020realtime} use motion characteristics to recognize divers, which would work neither in close proximity nor in the case of immobile divers. A possible solution would be to use facial cues for identification. In the literature, human subject identification using faces has been widely studied for terrestrial robots~\cite{hsu2018human, wang2019real, lee2020learning}. On the other hand, diver identification using faces has received little attention due to the unique challenges (see Section~\ref{intro}) present in the underwater domain.

In recent years, deep convolutional neural networks have yielded high-accuracies in face recognition tasks~\cite{taigman2014deepface, sun2014deep, schroff2015facenet}. In general, most of these methods include the following steps: 
\begin{enumerate}
\item \textbf{Face detection}: by learning feature pyramid~\cite{tang2018pyramidbox}, using a ``proposal and refinement'' mechanism~\cite{sun2018face}, or densely sampling locations across multiple scales~\cite{zhang2017single};
\item \textbf{Face classification}: achieved by either training a multi-class classifier to classify different identities in the training set using softmax loss~\cite{parkhi2015deep}, or directly learning the face representation (\ie embeddings) to classify different identities which may not be present in the training set~\cite{wen2016discriminative, sankaranarayanan2016triplet}. 
\end{enumerate}
However, it is often seen that the features learned using the softmax loss are not highly discriminative to represent a face outside of the training set~\cite{yan2019vargfacenet}. On the other hand, direct embedding learning methods show slower performance on large-scale datasets. Different solutions have been proposed~\cite{liu2017sphereface, wang2018cosface} to ensure compactness in \textit{intra-class distances} (\ie the same faces) while maximizing the \textit{inter-class distances} (\ie different faces).

Despite showing great promise for the general face recognition task, these methods exhibit poor performance on partially-occluded faces~\cite{yang2015robust}. Several methods have been proposed that extract local face descriptors only from the non-occluded facial areas~\cite{song2019occlusion}. In~\cite{elmahmudi2019deep}, a VGG-Face~\cite{parkhi2015deep} based feature extractor is used to classify faces from partial data. However, the discriminative power of these methods is limited. The pioneering work in occlusion-robust face recognition, proposed in~\cite{wright2009robust}, codes the occluded face image as a sparse linear combination of the training samples and the occlusion dictionary. The classification is then performed by minimizing the reconstruction loss with the sparse coding coefficients to recover the original faces from the occluded ones. Inspired by this approach, different variants have been proposed by changing the sparse constraint term~\cite{yang2011robust}; however, these methods fail to generalize because they assume test samples have identical subjects as the training samples. In~\cite{he2018dynamic}, a dynamic feature matching (DFM) technique is proposed where the training pictures are extracted as multiple sub-feature maps to match against the features of the probe face patches. In contrast,~\cite{singh2017disguised} propose a deep learning framework to detect facial keypoints from partial face images, and then the angles between the keypoints are used to perform face identification.

With the rapid evolution of generative adversarial networks (GAN)~\cite{goodfellow2014generative}, some researchers attempted to use generative methods to solve the partial face recognition problem. In~\cite{li2017generative}, first, different patches are corrupted (using Gaussian noise) in a face, and then GAN is used to reconstruct the occluded region to calculate similarity with the original non-occluded face. Zhao~\etal \cite{zhao2018robust} propose a long short term memory (LSTM)~\cite{hochreiter1997long} based autoencoder network~\cite{cheng2015robust} that can restore partially occluded faces. It consists of a multi-scale spatial LSTM encoder that can learn the latent representations by reading facial patches in various scales and a dual-channel recurrent LSTM decoder that decodes both occluded and non-occluded parts of faces to reconstruct the original faces. One of the potential use cases of these partial face recognition algorithms could be for an underwater diver identification system for an AUV, a research direction that is still unexplored. This research gap is mostly due to-
\begin{enumerate}
\item the difficulty in collecting real-world diver face datasets,
\item distortions present in underwater imagery~\cite{galdran2015automatic}, and
\item the complex facial occlusions occurring on diver faces. 
\end{enumerate}

\begin{figure*}[ht]
    \vspace{2mm}
    \centering
    \includegraphics[width=0.95\linewidth]{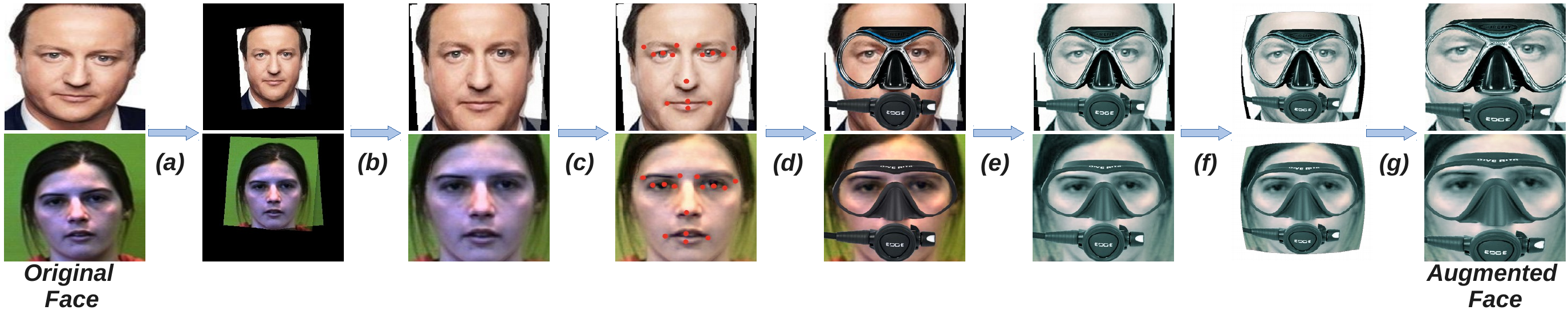}
    \caption{Data augmentation steps for preparing the training dataset for both the feature extraction and the diver face generation networks. (a) \textit{Frontalization}~\cite{hassner2015effective}. (b) Automatic crop. (c) Color correction and keypoints regression. (d) Apply mask and regulator based on the 15 keypoints. (e) Colorization~\cite{uwcorrect} to incorporate underwater lighting conditions. (f) Applying fish-eye effect~\cite{fisheye}. (g) Crop to make the boundary tight.}
    \label{fig:data_aug}
    \vspace{0mm}
\end{figure*}
\begin{figure}[t]
    \centering
    \begin{subfigure}{0.49\textwidth} 
    \centering
        \includegraphics[width=.90\linewidth]{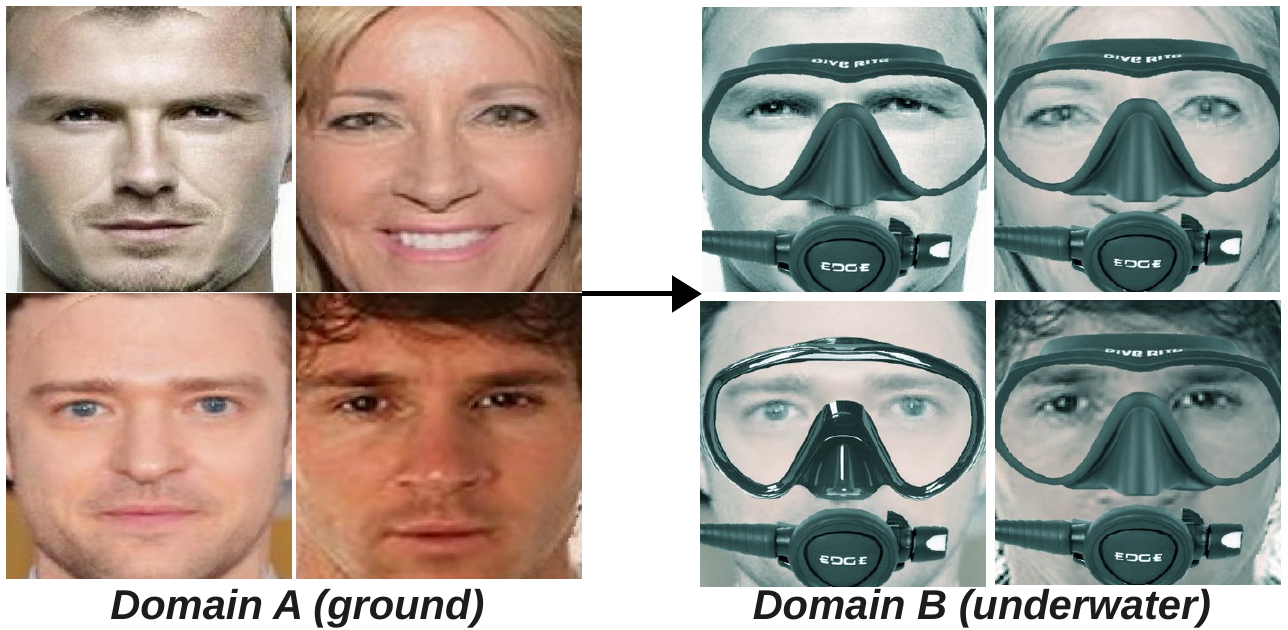}
        \caption{Patchwise contrastive learning for one-sided translation where $G_{enc}$ and $G_{dec}$ are sequentially applied to samples from domain $A$ to produce outputs that look like samples from domain $B$, \ie$\hat{y}_B=G_{dec}\Big(G_{enc}(x_A) \Big)$.}
        \label{fig:synthetic_faces1}
    \end{subfigure}
    \begin{subfigure}{0.48\textwidth} 
    \centering
        \includegraphics[width=.90\linewidth]{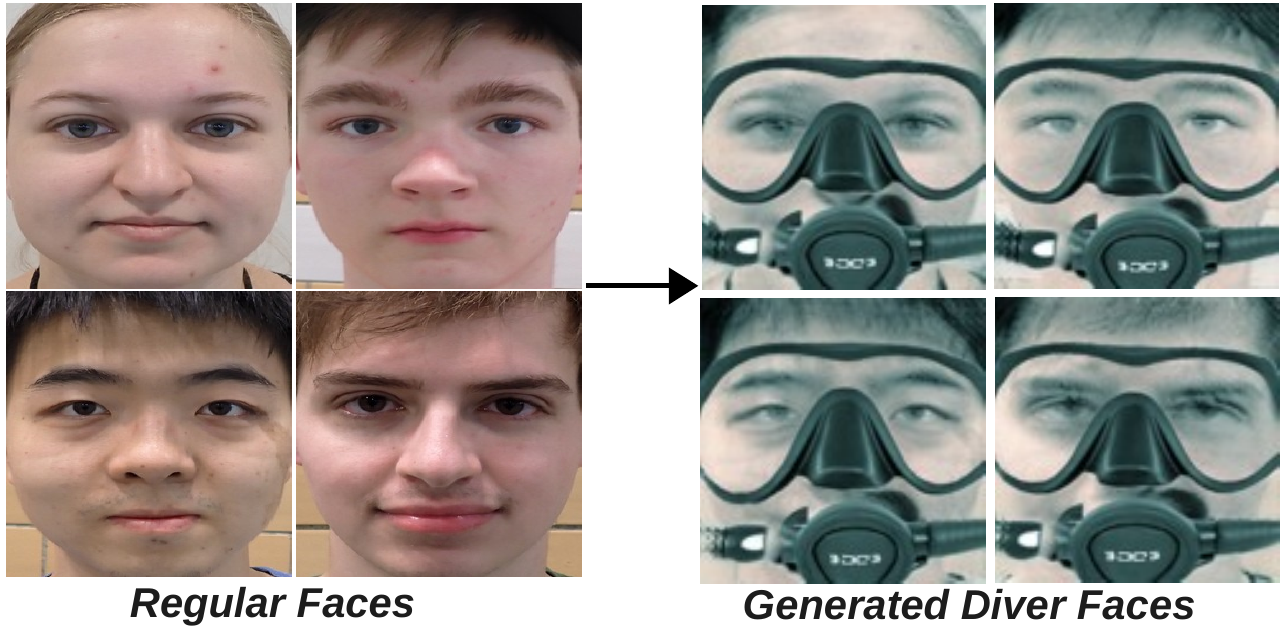}
        \caption{Converting the authorized users' regular faces into diver faces using the model learned in Fig.~\ref{fig:synthetic_faces1}. Feature embeddings are extracted from these faces and stored in a diver face identification database for real-time inference.}
        \label{fig:synthetic_faces2}
    \end{subfigure}
    \caption{Realistic diver face generation from regular face using an image-to-image translation method CUT~\cite{park2020contrastive}.}
    \label{fig:synthetic_faces}
    \vspace{0mm}
\end{figure}

\section{Methodology}
The overall pipeline of the proposed approach consists of both online and offline components. We train the feature extraction model completely offline to reduce the overhead during the inference time.

\subsection{Offline Module}

\subsubsection{Data Augmentation}\label{dataaug}
A significant challenge in the diver face recognition task is the unavailability of diver face datasets. To circumvent this issue, we resort to using publicly available non-underwater (\ie \textit{regular} or \textit{non-diver}) face datasets, and perform a series of operations, described below, to transform them into \textit{diver's} faces. Our \textit{augmented} face dataset thus has both regular (non-diver) and diver face pairs. Firstly, we perform \textit{frontalization} to the faces in our dataset, as described in~\cite{hassner2015effective}. That is, given a \textit{query} photo, we detect facial features on both the query photo and a rendered 3D face model. The 2D coordinates on the query image and the corresponding 3D coordinates on the 3D model enable us to estimate a projection matrix to project the query points onto a reference coordinate system. Finally, the missing pixels' color intensities are filled in with the original query image's symmetric regions. We perform an automatic crop on the processed image to retrieve the original composition of the image.

Next, we use a facial keypoint prediction network, which we have constructed using 4 convolutional layers and 3 fully connected layers, that outputs 30 $(x,y)$ locations (15 keypoint coordinates). These 15 keypoints are used to put different kinds of masks and snorkels on top of the faces to make them look like diver faces. We process the masked images further~\cite{uwcorrect} to achieve the underwater green/blue hue. In addition, to account for distortions (\eg barrel distortions~\cite{schwalbe2005geometric}) caused by underwater optics, our training face images need to mimic the appearance of faces coming from underwater robot cameras. We achieve this by adding fish-eye effects~\cite{fisheye} to the face images in our dataset. Fig.~\ref{fig:data_aug} demonstrates all the steps of the data augmentation process.


\subsubsection{Feature Extraction Method} \label{feat_extract_method}
The main challenge in a face recognition task is to learn discriminative features that maximize inter-class distances and minimize intra-class distances. We select ArcFace~\cite{deng2019arcface} as a baseline model to extract features since it has shown the highest accuracy among the recent SOTA algorithms. During the training, we utilized the following techniques to improve the feature extraction: center cropping of images, \textit{frontalizing} faces, varying the size of the embeddings and changing the composition of the training dataset (\ie diver + non-diver faces, diver faces only, etc.). Essentially, the network learns to obtain a highly discriminative 512-dimensional feature embedding to represent a diver face.

\subsubsection{Diver Face Generation} \label{face_gen}
As explained in Section~\ref{dataaug}, it takes several steps to create realistic diver faces from non-diver faces, which could be laborious when adding extra divers to an existing database during an ongoing mission. To avoid the data pre-processing delays, we use generative models, specifically image-to-image translation~\cite{isola2017image} techniques, to generate underwater faces from terrestrial ones. There are two types of image-to-image translation methods: 
\begin{enumerate}
\item Supervised models with paired training data, and 
\item Unsupervised models with unpaired training data.
\end{enumerate}
Although our data augmentation process provides us with a paired dataset, we intentionally use an unsupervised model for the following reasons: filtering augmented images could break pairs, and it is easier to expand the training data by relaxing the constraint of requiring image pairs.

Among unsupervised image-to-image translation models (\eg\cite{zhu2017unpaired, kim2017learning, choi2020stargan}), we generate diver faces using CUT (Contrastive Unpaired Translation)~\cite{park2020contrastive} since it achieves a superior Fr\`echet Inception Distance (FID) and mean average precision (mAP) scores compared to the models listed above. The FID score represents the statistical similarity between generated and real images, which is especially important for our use case.  Fig.~\ref{fig:synthetic_faces} shows the diver face generation process.

\subsubsection{Diver Face Identification Database}
We create a diver face identification database which stores the 512-dimensional face embeddings of the authorized users. These embeddings are extracted from the corresponding generated diver faces, using the method described in Section~\ref{face_gen}, from the regular (non-diver) faces of the authorized users who are allowed to control the AUV. The embeddings are extracted using the trained model, described in Section~\ref{feat_extract_method}. We store at least 2 extracted embeddings for each subject to minimize the false negative rate.

\subsection{Online Module}

\subsubsection{Diver Face Localization}
In the M/HRI context, an AUV needs to authenticate a diver before they are authorized to issue commands. First, the AUV needs to localize or detect diver faces. To this end, we found that RetinaFace~\cite{deng2019retinaface} performed best in localizing diver faces in underwater images compared to other face detection algorithms (\eg \cite{li19dual,zhang16joint, zhang2019improved}). RetinaFace is a single-stage pixel-wise face localization method which employs a multi-task learning strategy to detect faces. The algorithm uses an extra-supervised branch in regressing five facial landmarks and a self-supervised branch in predicting 3D position and correspondence of each facial pixel on top of the typical face score and face box prediction branch.

\subsubsection{Diver Face Matching} \label{face_matching}
\begin{figure}[ht]
\vspace{2mm}
    \centering
    \includegraphics[width=0.99\linewidth]{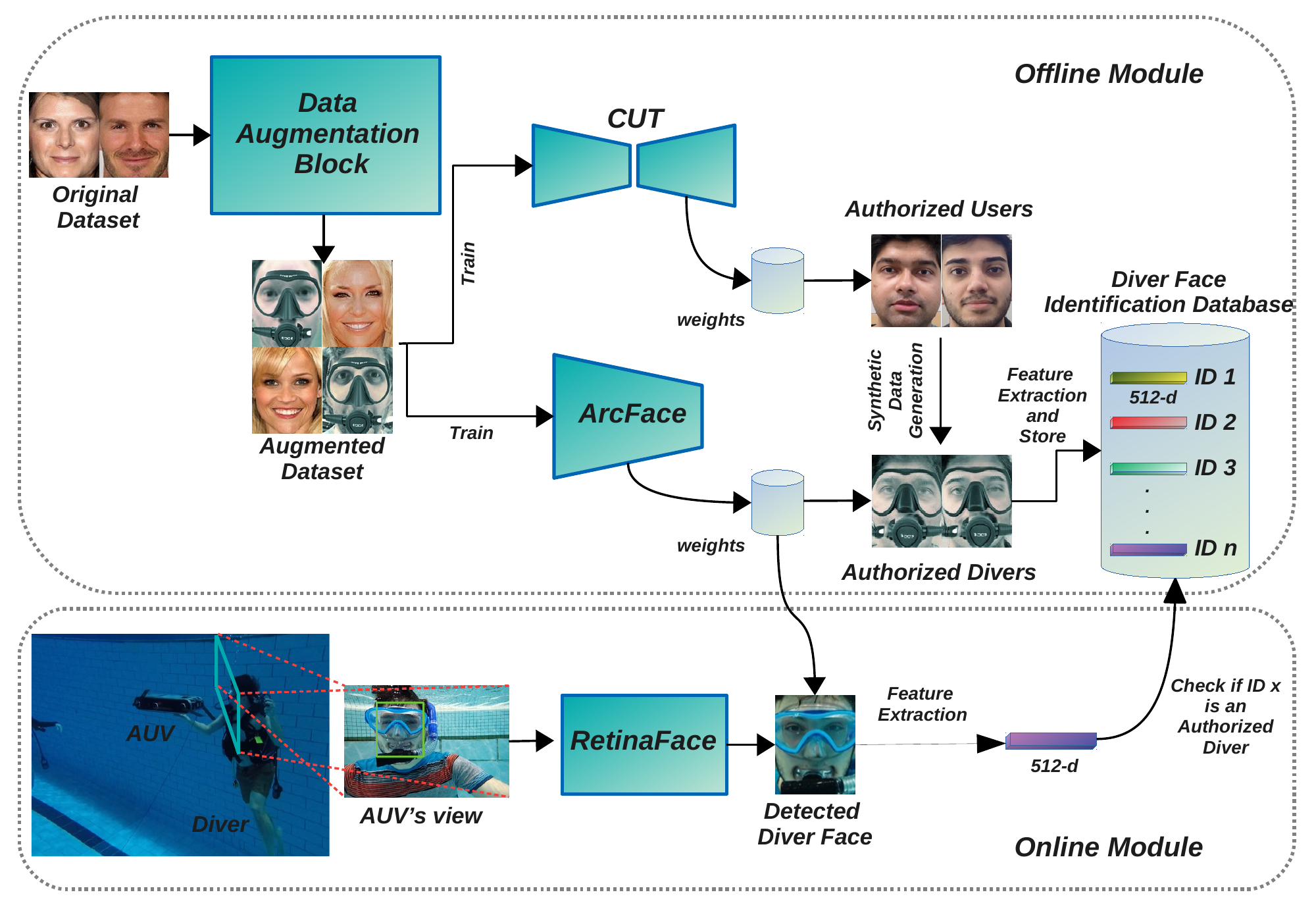}
    \caption{Complete pipeline of the proposed diver face identification system. The overall system consists of an online as well as an offline component. In real-world scenarios, only the online module is executed by AUVs to recognize scuba divers, \eg to ensure they are allowed to operate the AUVs.}
    \label{fig:complete_pipeline}
\end{figure}

Once the AUV has detected a diver face from the captured image, it will extract highly-robust discriminative feature embedding from the diver face using the method described in Section~\ref{feat_extract_method}. Once this \textit{query} embedding is calculated, the algorithm will try to match it against all the authorized users' embeddings stored in the database. If a match is found, the diver would be granted access to operate the AUV. There are 2 other alternative matching options available:
\begin{enumerate}
\item Match regular (non-diver) faces with diver faces, and  
\item Match regular (non-diver) faces with converted diver faces (to non-diver faces) using CUT.
\end{enumerate}
These options eliminate most of the data augmentation steps; however, they perform poorly, as discussed in Section~\ref{results}. Fig.~\ref{fig:complete_pipeline} shows the complete pipeline of the proposed diver face identification framework. 

\begin{figure}[t]
    \vspace{2mm}
    \centering
    \includegraphics[width=0.99\linewidth]{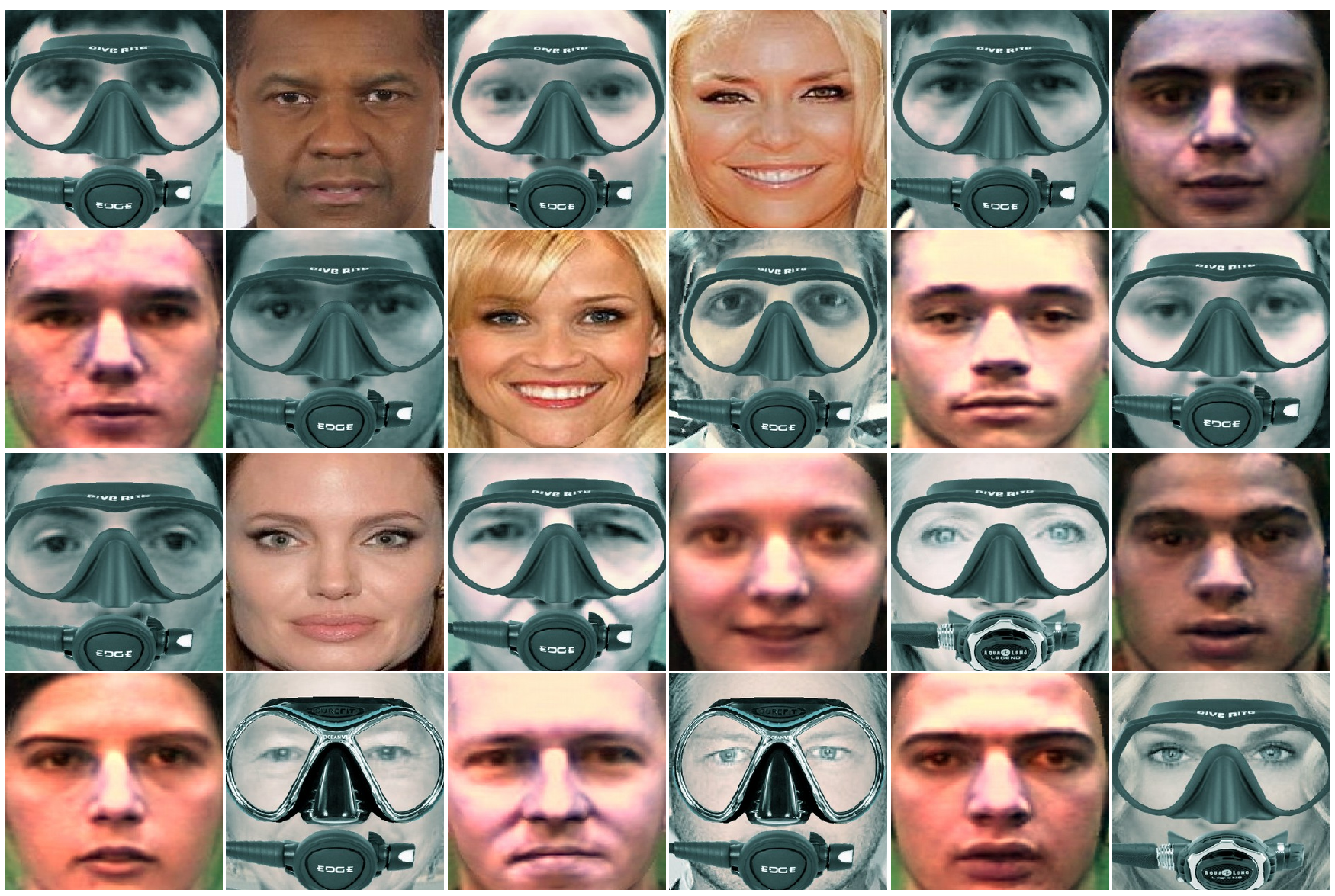}
    \caption{A few instances sampled from the proposed dataset. The final prepared dataset includes a total number of $22,031$ (diver + non-diver) face images with 184 different subjects.}
    \label{fig:dataset}
    \vspace{0mm}
\end{figure}
\begin{table}[t]
\vspace{2mm}
\centering
\caption{Setup for each network.}
\footnotesize
\begin{tabular}{|l||c|c|c|c|c}
  \hline
  \textbf{Model} & Epochs  & GPU & Library & Pretrained  \\ \hline \hline
  Keypoint &  - & - & Tensorflow~\cite{abadi2016tensorflow} & \checkmark \\
  \hline
  RetinaFace & - & - & Tensorflow & \checkmark \\
  \hline
  ArcFace & 3000 &  2080 Ti & Tensorflow & - \\
  \hline
  FaceNet & - &  - & Tensorflow & \checkmark  \\
  \hline
  CUT & 500 &  Titan X & PyTorch~\cite{paszke2019pytorch} & - \\
  \hline
\end{tabular}
\label{tab:exp}
\end{table}%

\section{Experimental Setup and Evaluation}
\subsection{Dataset} \label{sec:dataset}
There are $184$ different subjects in the dataset, collected from the following sources~\cite{sengupta16frontal, faces94}. Each subject has multiple regular face images. For our purpose, we augment the dataset by putting $4$ different kinds of masks and $2$ different kinds of snorkels on top of the regular faces. This gives us a total number of $22,031$ (diver + non-diver) samples of $184$ different subjects. We use different combinations of masks and snorkels so that our feature extraction model learns generalized discriminative face features. Finally, we carry out the steps described in Section~\ref{dataaug} to prepare the final dataset. We show a few sample images from the prepared dataset in Fig.~\ref{fig:dataset}.

\begin{table*}[t]
\vspace{2mm}
\centering
\caption{Diver face identification performance of different methods on real-world data.}
\footnotesize
\begin{tabular}{|l||c|c|c|c|c|}
  \hline
  \textbf{Approach} &Training Data & Backbone & \textit{Frontalization} & Embedding Size &  Prediction Accuracy $(\%)$ \\ \hline \hline
  FaceNet~\cite{schroff2015facenet} & -&- & - & 128 & 16.67   \\
  \hline
  ArcFace~\cite{deng2019arcface} &-& ResNet-50~\cite{he2016deep} & - & 512 & 16.67   \\
  \hline
  Variant 1 & diver + non-diver & ResNet-50 &no & 512& 16.67   \\
  \hline
  Variant 2 & diver + non-diver & ResNet-50 & no & 512 & 16.67   \\
  \hline
  Variant 3 & diver & ResNet-50 & yes & 512 & 50.00  \\
  \hline
  Variant 4 & diver & ResNet-50 & yes & 256 & 33.33  \\
  \hline
  Variant 5 & diver & ResNet-50 & yes & 1024 & 33.33   \\
  \hline
  Ours & diver & ResNet-50 & yes & 512 & \textbf{66.67} \\ \hline
\end{tabular}
\vspace{-4mm}
\label{tab:quant_eval}
\end{table*}%

\begin{figure*}[t]
\vspace{2mm}
    \centering
    \begin{subfigure}{0.50\textwidth} 
    \centering
        \includegraphics[width=1\linewidth]{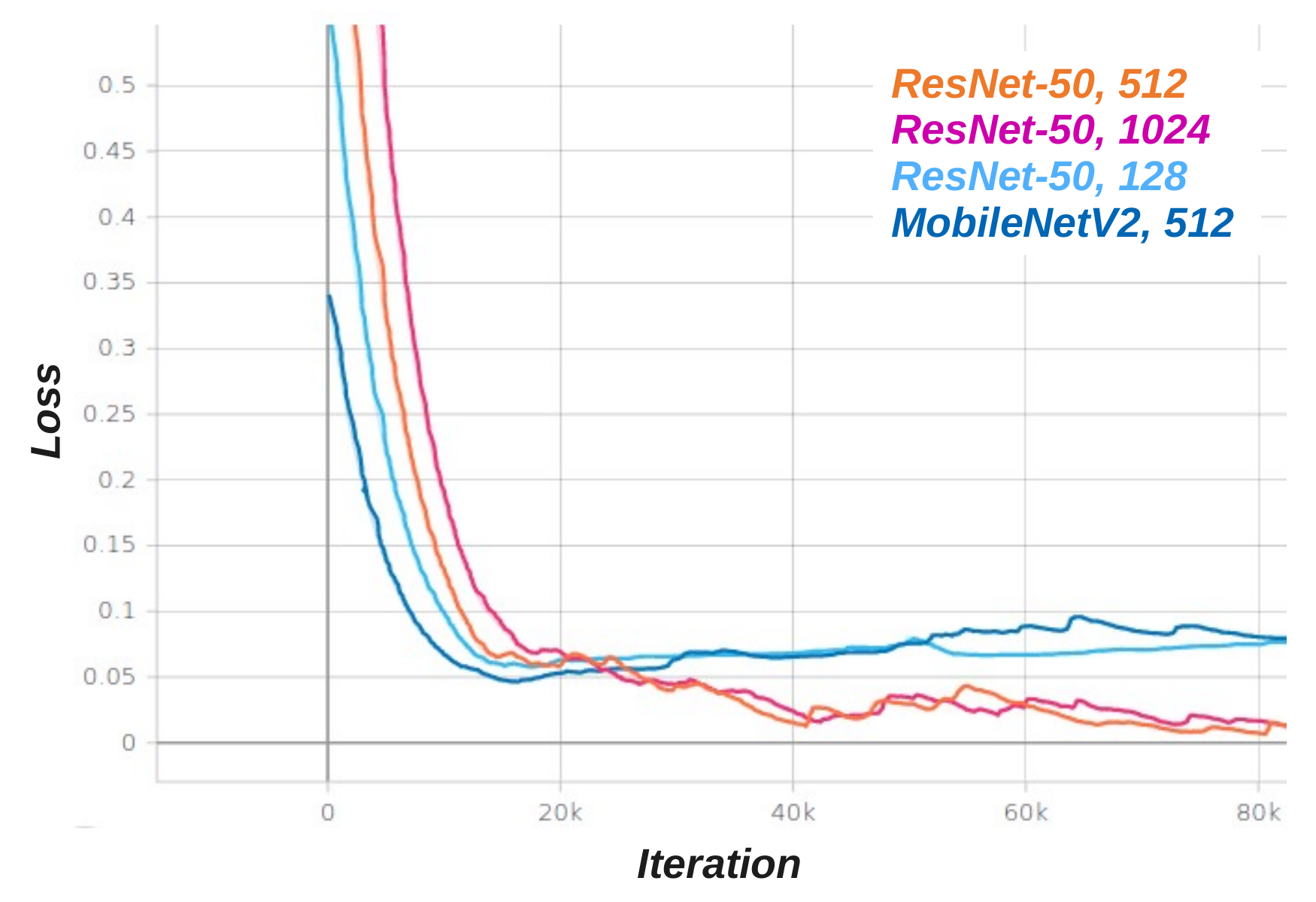}
        \caption{Validation loss vs number of iterations for the feature extraction network as we change the backbone of the network and the size of the extracted feature embedding. Here, ``Backbone, $X$" denotes different backbones that are being used while $X=$ embedding size.}
        \label{fig:backbone1}
    \end{subfigure}
    \vspace{2mm}
    \begin{subfigure}{0.43\textwidth} 
    \centering
        \includegraphics[width=.85\linewidth]{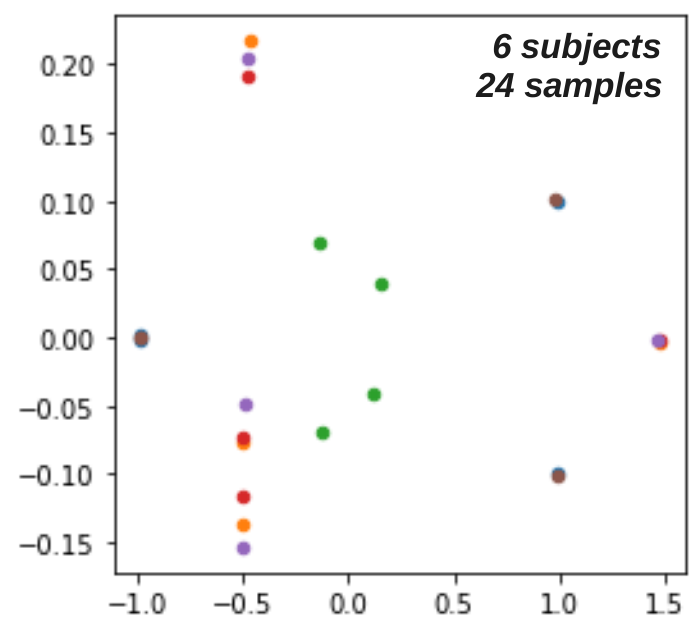}
        \caption{512-dimensional features (learnt using MobileNetV2 as the backbone) are plotted in a 2-dimensional space by reducing their dimensions using PCA. Different color represents different subjects.}
        \label{fig:backbone2}
    \end{subfigure}
    \caption{Performance of the feature extraction network with different backbones and feature embedding sizes on real-world data.}
    \label{fig:backbone}
\vspace{0mm}
\end{figure*}

\begin{figure}[t]
\vspace{2mm}
    \centering
    \includegraphics[width=0.6\linewidth]{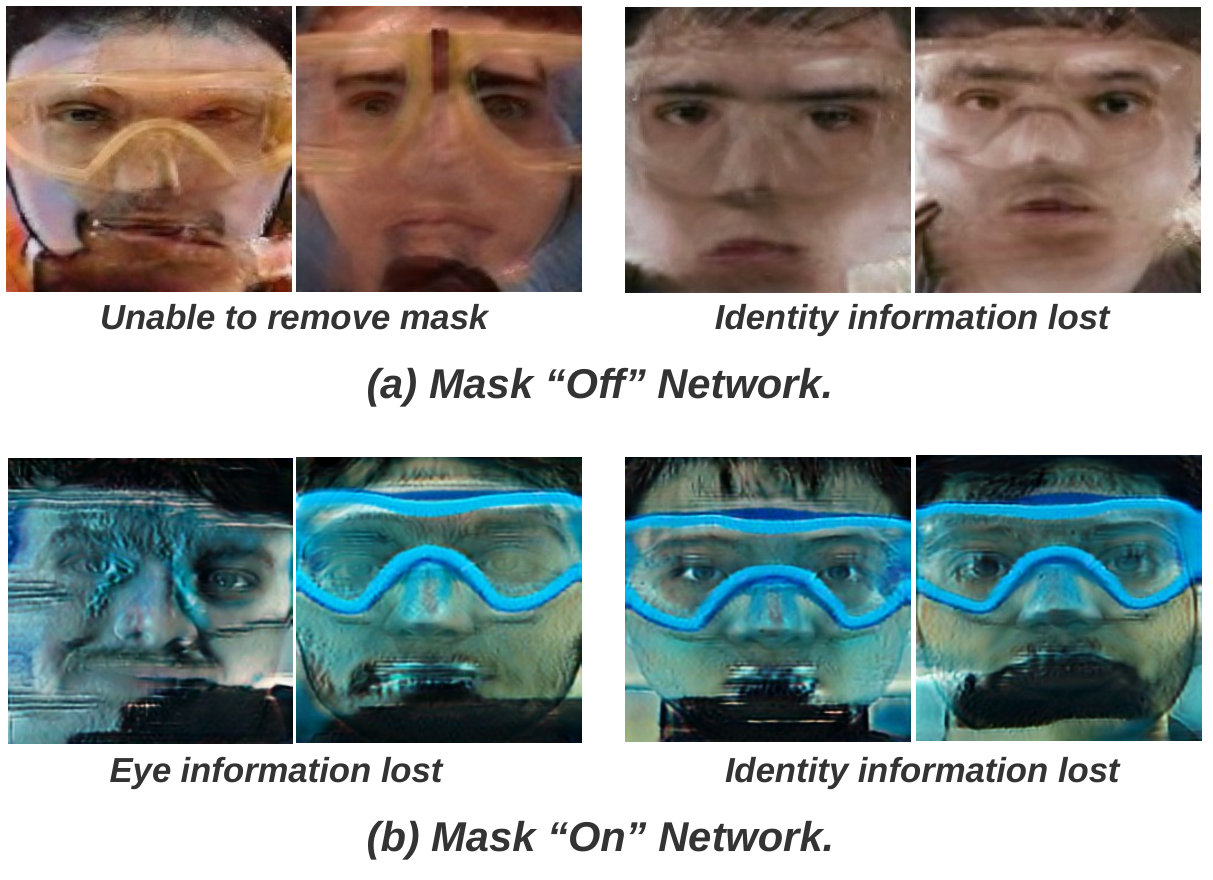}
    \caption{Few failure cases of the proposed framework. (a) It tends to struggle while trying to remove masks from the diver faces. (b) Improperly trained data generation model completely destroys the underlying information of the image.}
    \label{fig:failure_cases}
\end{figure}

\subsection{Setup}
Table \ref{tab:exp} shows our experimental setup. We used pre-trained weights if trained on large datasets (\eg WIDER FACE dataset~\cite{yang2016wider}). Both ArcFace and CUT are trained with our augmented dataset (see Section~\ref{sec:dataset}). 

Additionally, we run inference with RetinaFace and ArcFace on Nvidia Jetson Nano to test prediction latency on a mobile GPU platform commonly used in AUVs.   

\subsection{Evaluation Criteria} \label{criteria}
We evaluate the performance of the proposed approach on real-world data by matching the diver faces against the stored faces in our database. First, the algorithm extracts a 512-dimensional feature embedding from a detected scuba diver's face. Then this \textit{query} embedding is compared against the stored embeddings of the \textit{authorized} users by calculating the cosine similarity (CS)~\cite{nguyen2010cosine} between them. The CS between two feature vectors $f_q$ (embedding of the query face) and $f_a$ (embedding of the authorized face) is defined as, 
\begin{equation*}
    CS(f_q,f_a)=\frac{f_q^Tf_a}{\Vert f_q \Vert \Vert f_a \Vert }
\end{equation*}

If the CS between the query embedding and a specific stored embedding is the maximum, we check if they are the \textit{same} subject. If they are the same, then the prediction is considered correct. To calculate the accuracy of the predictions, we take some test samples (\ie real-world diver faces) to match them against the stored subjects and keep a count of the correct predictions. Finally, we divide this number by the total number of test samples to get the final prediction accuracy.

\subsection{Evaluation Results} \label{results}
To test the performance of the proposed framework, we select 6 scuba divers as our test subjects. First, we capture their regular (non-diver, \ie not wearing any breathing apparatus) faces on the ground. These ground pictures are converted to diver faces following the approach in Section~\ref{face_gen}, and embeddings are extracted and stored in the diver face identification database. Then, the faces of the divers are captured underwater from which \textit{query} embeddings are extracted. Using the evaluation criteria described in Section~\ref{criteria}, we calculate the prediction accuracy of the framework for the 6 subjects. In a similar fashion, we calculate the prediction accuracy of a few other SOTA face recognition algorithms. Further, we document the performances of 5 different variants of our framework, which are created by varying the composition of the training dataset, adding/removing \textit{frontalization}, varying the size of the embeddings, as described in Section~\ref{feat_extract_method}. The performances are listed in Table~\ref{tab:quant_eval}. As can be seen from the table, our method achieves a superior prediction accuracy of $66.67\%$, compared to other algorithms. Even the SOTA algorithms~\cite{deng2019arcface} struggle to learn robust discriminative features from diver faces and subsequently perform poorly in the diver face recognition task.

Moreover, we experiment with different backbones for our feature extraction network. Fig.~\ref{fig:backbone} shows the performance of the feature extraction model when trained with only diver face images. From Fig.~\ref{fig:backbone1} we see, \textit{ResNet-50, 512} and \textit{ResNet-50, 1024} tend to converge after 20k iterations. However, if we decrease the size of the embeddings (\eg 128), the network starts to diverge after some time, indicating that it is no longer capable of learning discriminative features from the data. A similar diverging pattern exists if we use MobileNetV2~\cite{sandler2018mobilenetv2} as the backbone of the network regardless of the size of the feature embedding. To investigate further, we try to see where the feature embeddings of different subject classes lie in a Euclidean space if MobileNetV2 is used. For visualization purpose, we perform dimensionality reduction using Principal Component Analysis (PCA)~\cite{wold1987principal} on 24 extracted feature embeddings of 6 subjects (4 embeddings/subject) and project them in a 2-dimensional space, as shown in Fig.~\ref{fig:backbone2}. From the figure, it is evident that the class boundaries for the subjects are highly overlapping except for the one in green. Similar results follow even if we use ResNet-50 with 128-dimensional feature embeddings. Therefore, we decide to use 512-dimensional embeddings, with ResNet-50 as the backbone.  

Next, we evaluate the performance of different similarity matching approaches, as described in Section~\ref{face_matching}. We find that none of the algorithms translate well if we try to match unprocessed regular (non-diver) faces with real-world diver faces. Also, we convert the real-world diver faces into non-diver faces (\ie take the mask off) using CUT, but this process either fails to remove the mask from the divers' faces or completely destroys the identity of the users as seen in Fig.~\ref{fig:failure_cases}a. Therefore, we decide to match processed non-diver faces (\ie generated diver faces) and real-world diver faces. However, there are a few instances where our approach fails. As shown in Fig.~\ref{fig:failure_cases}b, we found that an improperly trained data generation network fails to convert the non-diver faces into diver faces, which reduces the prediction accuracy. 

\subsection{Practical Feasibility}
The accuracy of our model is not as high as the SOTA non-diver face recognition models, which achieve higher than 97+\%~\cite{masi2018deep} on the LFW dataset~\cite{huang2007labeled}. However, our diver face prediction accuracy is around 3-times higher than the SOTA face recognition algorithms, as shown in Table~\ref{tab:quant_eval}. Overall our recognition results show that our approach is effective in capturing discriminative features from diver faces.

On our testing on an Nvidia Jetson Nano embedded GPU, inference took $\approx 36$ \textit{ms} while loading the models required $\approx 2$ \textit{minutes}. The results lead us to believe that the models can run in real-time. However, the loading time needs to be improved by reducing the model size in case of multiple deep learning models are mounted on a robot.

\section{Conclusions and Future Work}
This paper presents our work on a diver face
recognition system using data augmentation techniques and
generative models. The results show that our proposed system can outperform the SOTA face recognition algorithms, and the models could be deployed on embedded GPU hardware commonly found in real robotic platforms. Our approach \textit{(i)} allows secure human-robot collaboration without any extra devices other than visual sensors, \textit{(ii)} does not need any prior knowledge about divers other than their non-diver headshot photos, and \textit{(iii)} can achieve reasonable accuracy without using pairs of diver and non-diver face images from each subject during training. We intend to improve this work in several directions. We are working on improving our \textit{Mask ``Off"} generative model to take advantage of existing face recognition models. Also, we plan to collect more diver faces from various underwater environments to develop a more robust system. Lastly, we are working on reducing the model's size so a robot can run our system along with other deep learning modules.

\section*{Acknowledgement}
We gratefully acknowledge the GPU grants from Nvidia Corporation that made this research possible. 


\bibliographystyle{IEEEtran}
\bibliography{ref.bib}

\begin{thebibliography}{10}
\providecommand{\url}[1]{#1}
\csname url@rmstyle\endcsname
\providecommand{\newblock}{\relax}
\providecommand{\bibinfo}[2]{#2}
\providecommand\BIBentrySTDinterwordspacing{\spaceskip=0pt\relax}
\providecommand\BIBentryALTinterwordstretchfactor{4}
\providecommand\BIBentryALTinterwordspacing{\spaceskip=\fontdimen2\font plus
\BIBentryALTinterwordstretchfactor\fontdimen3\font minus
  \fontdimen4\font\relax}
\providecommand\BIBforeignlanguage[2]{{%
\expandafter\ifx\csname l@#1\endcsname\relax
\typeout{** WARNING: IEEEtran.bst: No hyphenation pattern has been}%
\typeout{** loaded for the language `#1'. Using the pattern for}%
\typeout{** the default language instead.}%
\else
\language=\csname l@#1\endcsname
\fi
#2}}

\bibitem{xanthidis2020navigation}
M.~Xanthidis, N.~Karapetyan, H.~Damron, S.~Rahman, J.~Johnson, A.~O’Connell,
  J.~M. O’Kane, and I.~Rekleitis, ``{Navigation in the Presence of Obstacles
  for an Agile Autonomous Underwater Vehicle},'' in \emph{2020 IEEE
  International Conference on Robotics and Automation (ICRA)}.\hskip 1em plus
  0.5em minus 0.4em\relax IEEE, 2020, pp. 892--899.

\bibitem{manderson2020vision}
T.~Manderson, J.~C.~G. Higuera, S.~Wapnick, J.-F. Tremblay, F.~Shkurti,
  D.~Meger, and G.~Dudek, ``{Vision-Based Goal-Conditioned Policies for
  Underwater Navigation in the Presence of Obstacles},'' in \emph{Robotics:
  Science and Systems}, Corvalis, Oregon, USA, 2020.

\bibitem{fulton2019robotic}
M.~Fulton, J.~Hong, M.~J. Islam, and J.~Sattar, ``{Robotic Detection of Marine
  Litter Using Deep Visual Detection Models},'' in \emph{2019 International
  Conference on Robotics and Automation (ICRA)}.\hskip 1em plus 0.5em minus
  0.4em\relax IEEE, 2019, pp. 5752--5758.

\bibitem{islam2018dynamic}
M.~J. {Islam}, M.~{Ho}, and J.~{Sattar}, ``{Dynamic Reconfiguration of Mission
  Parameters in Underwater Human-Robot Collaboration},'' in \emph{2018 IEEE
  International Conference on Robotics and Automation (ICRA)}, 2018, pp.
  6212--6219.

\bibitem{xia2019visual}
Y.~Xia and J.~Sattar, ``{Visual Diver Recognition for Underwater Human-Robot
  Collaboration},'' in \emph{2019 International Conference on Robotics and
  Automation (ICRA)}.\hskip 1em plus 0.5em minus 0.4em\relax IEEE, 2019, pp.
  6839--6845.

\bibitem{de2020realtime}
K.~de~Langis and J.~Sattar, ``{Realtime Multi-Diver Tracking and
  Re-identification for Underwater Human-Robot Collaboration},'' in \emph{2020
  IEEE International Conference on Robotics and Automation (ICRA)}.\hskip 1em
  plus 0.5em minus 0.4em\relax IEEE, 2020, pp. 11\,140--11\,146.

\bibitem{islam2019understanding}
M.~J. Islam, M.~Ho, and J.~Sattar, ``{Understanding human motion and gestures
  for underwater human-robot collaboration},'' \emph{Journal of Field
  Robotics}, vol.~36, no.~5, pp. 851--873, 2019.

\bibitem{weidner2017underwater}
N.~Weidner, S.~Rahman, A.~Q. Li, and I.~Rekleitis, ``{Underwater Cave Mapping
  using Stereo Vision},'' in \emph{2017 IEEE International Conference on
  Robotics and Automation (ICRA)}.\hskip 1em plus 0.5em minus 0.4em\relax IEEE,
  2017, pp. 5709--5715.

\bibitem{islam2020sesr}
M.~J. Islam, P.~Luo, and J.~Sattar, ``{Simultaneous Enhancement and
  Super-Resolution of Underwater Imagery for Improved Visual Perception},'' in
  \emph{Robotics: Science and Systems (RSS)}, 2020.

\bibitem{deng2019arcface}
J.~{Deng}, J.~{Guo}, N.~{Xue}, and S.~{Zafeiriou}, ``{ArcFace: Additive Angular
  Margin Loss for Deep Face Recognition},'' in \emph{2019 IEEE/CVF Conference
  on Computer Vision and Pattern Recognition (CVPR)}, 2019, pp. 4685--4694.

\bibitem{wang2018cosface}
H.~{Wang}, Y.~{Wang}, Z.~{Zhou}, X.~{Ji}, D.~{Gong}, J.~{Zhou}, Z.~{Li}, and
  W.~{Liu}, ``{CosFace: Large Margin Cosine Loss for Deep Face Recognition},''
  in \emph{2018 IEEE/CVF Conference on Computer Vision and Pattern
  Recognition}, 2018, pp. 5265--5274.

\bibitem{UnderwaterOptics}
T.~T. Team, ``{The Physics of Diving: Light and Vision},''
  \url{http://library.thinkquest.org/28170/35.html}.

\bibitem{dudek2007visual}
G.~Dudek, J.~Sattar, and A.~Xu, ``{A Visual Language for Robot Control and
  Programming: A Human-Interface Study},'' in \emph{Proceedings 2007 IEEE
  International Conference on Robotics and Automation}.\hskip 1em plus 0.5em
  minus 0.4em\relax IEEE, 2007, pp. 2507--2513.

\bibitem{verzijlenberg2010swimming}
B.~Verzijlenberg and M.~Jenkin, ``{Swimming with Robots: Human Robot
  Communication at Depth},'' in \emph{2010 IEEE/RSJ International Conference on
  Intelligent Robots and Systems}.\hskip 1em plus 0.5em minus 0.4em\relax IEEE,
  2010, pp. 4023--4028.

\bibitem{hsu2018human}
S.-C. Hsu, Y.-W. Wang, and C.-L. Huang, ``{Human Object Identification for
  Human-Robot Interaction by using Fast R-CNN},'' in \emph{2018 Second IEEE
  International Conference on Robotic Computing (IRC)}.\hskip 1em plus 0.5em
  minus 0.4em\relax IEEE, 2018, pp. 201--204.

\bibitem{wang2019real}
Y.~Wang, J.~Shen, S.~Petridis, and M.~Pantic, ``{A real-time and unsupervised
  face Re-Identification system for Human-Robot Interaction},'' \emph{Pattern
  Recognition Letters}, vol. 128, pp. 559--568, 2019.

\bibitem{lee2020learning}
Y.~L. {Lee}, M.~Y. {Tseng}, Y.~C. {Luo}, D.~R. {Yu}, and W.~C. {Chiu},
  ``{Learning Face Recognition Unsupervisedly by Disentanglement and
  Self-Augmentation},'' in \emph{2020 IEEE International Conference on Robotics
  and Automation (ICRA)}, 2020, pp. 3018--3024.

\bibitem{taigman2014deepface}
Y.~{Taigman}, M.~{Yang}, M.~{Ranzato}, and L.~{Wolf}, ``{DeepFace: Closing the
  Gap to Human-Level Performance in Face Verification},'' in \emph{2014 IEEE
  Conference on Computer Vision and Pattern Recognition}, 2014, pp. 1701--1708.

\bibitem{sun2014deep}
Y.~Sun, Y.~Chen, X.~Wang, and X.~Tang, ``{Deep Learning Face Representation by
  Joint Identification-Verification},'' in \emph{Advances in Neural Information
  Processing Systems}.\hskip 1em plus 0.5em minus 0.4em\relax Curran
  Associates, Inc., 2014, pp. 1988--1996.

\bibitem{schroff2015facenet}
F.~{Schroff}, D.~{Kalenichenko}, and J.~{Philbin}, ``{FaceNet: A Unified
  Embedding for Face Recognition and Clustering},'' in \emph{2015 IEEE
  Conference on Computer Vision and Pattern Recognition (CVPR)}, 2015, pp.
  815--823.

\bibitem{tang2018pyramidbox}
X.~Tang, D.~K. Du, Z.~He, and J.~Liu, ``{PyramidBox: A Context-assisted Single
  Shot Face Detector},'' in \emph{European Conference on Computer Vision
  (ECCV)}, 2018.

\bibitem{sun2018face}
X.~Sun, P.~Wu, and S.~C. Hoi, ``{Face detection using deep learning: An
  improved faster RCNN approach},'' \emph{Neurocomputing}, vol. 299, pp.
  42--50, 2018.

\bibitem{zhang2017single}
S.~{Zhang}, X.~{Zhu}, Z.~{Lei}, H.~{Shi}, X.~{Wang}, and S.~Z. {Li},
  ``{S$^3$FD: Single Shot Scale-Invariant Face Detector},'' in \emph{2017 IEEE
  International Conference on Computer Vision (ICCV)}, 2017, pp. 192--201.

\bibitem{parkhi2015deep}
O.~M. Parkhi, A.~Vedaldi, and A.~Zisserman, ``{Deep Face Recognition},'' in
  \emph{British Machine Vision Conference (BMVC)}.\hskip 1em plus 0.5em minus
  0.4em\relax BMVA Press, 2015, pp. 41.1--41.12.

\bibitem{wen2016discriminative}
Y.~Wen, K.~Zhang, Z.~Li, and Y.~Qiao, ``{A Discriminative Feature Learning
  Approach for Deep Face Recognition},'' in \emph{European Conference on
  Computer Vision}.\hskip 1em plus 0.5em minus 0.4em\relax Springer, 2016, pp.
  499--515.

\bibitem{sankaranarayanan2016triplet}
S.~{Sankaranarayanan}, A.~{Alavi}, C.~D. {Castillo}, and R.~{Chellappa},
  ``{Triplet Probabilistic Embedding for Face Verification and Clustering},''
  in \emph{2016 IEEE 8th International Conference on Biometrics Theory,
  Applications and Systems (BTAS)}, 2016, pp. 1--8.

\bibitem{yan2019vargfacenet}
M.~{Yan}, M.~{Zhao}, Z.~{Xu}, Q.~{Zhang}, G.~{Wang}, and Z.~{Su},
  ``{VarGFaceNet: An Efficient Variable Group Convolutional Neural Network for
  Lightweight Face Recognition},'' in \emph{2019 IEEE/CVF International
  Conference on Computer Vision Workshop (ICCVW)}, 2019, pp. 2647--2654.

\bibitem{liu2017sphereface}
W.~{Liu}, Y.~{Wen}, Z.~{Yu}, M.~{Li}, B.~{Raj}, and L.~{Song}, ``{SphereFace:
  Deep Hypersphere Embedding for Face Recognition},'' in \emph{2017 IEEE
  Conference on Computer Vision and Pattern Recognition (CVPR)}, 2017, pp.
  6738--6746.

\bibitem{yang2015robust}
H.~Yang, X.~He, X.~Jia, and I.~Patras, ``{Robust face alignment under occlusion
  via regional predictive power estimation},'' \emph{IEEE Transactions on Image
  Processing}, vol.~24, no.~8, pp. 2393--2403, 2015.

\bibitem{song2019occlusion}
L.~{Song}, D.~{Gong}, Z.~{Li}, C.~{Liu}, and W.~{Liu}, ``{Occlusion Robust Face
  Recognition Based on Mask Learning With Pairwise Differential Siamese
  Network},'' in \emph{2019 IEEE/CVF International Conference on Computer
  Vision (ICCV)}, 2019, pp. 773--782.

\bibitem{elmahmudi2019deep}
A.~Elmahmudi and H.~Ugail, ``{Deep face recognition using imperfect facial
  data},'' \emph{Future Generation Computer Systems}, vol.~99, pp. 213--225,
  2019.

\bibitem{wright2009robust}
J.~{Wright}, A.~Y. {Yang}, A.~{Ganesh}, S.~S. {Sastry}, and Y.~{Ma}, ``{Robust
  Face Recognition via Sparse Representation},'' \emph{IEEE Transactions on
  Pattern Analysis and Machine Intelligence}, vol.~31, no.~2, pp. 210--227,
  2009.

\bibitem{yang2011robust}
M.~{Yang}, L.~{Zhang}, J.~{Yang}, and D.~{Zhang}, ``{Robust Sparse Coding for
  Face Recognition},'' in \emph{CVPR 2011}, 2011, pp. 625--632.

\bibitem{he2018dynamic}
L.~He, H.~Li, Q.~Zhang, and Z.~Sun, ``{Dynamic Feature Learning for Partial
  Face Recognition},'' in \emph{2018 IEEE/CVF Conference on Computer Vision and
  Pattern Recognition}, 2018, pp. 7054--7063.

\bibitem{singh2017disguised}
A.~{Singh}, D.~{Patil}, G.~M. {Reddy}, and S.~N. {Omkar}, ``{Disguised Face
  Identification (DFI) with Facial KeyPoints Using Spatial Fusion Convolutional
  Network},'' in \emph{2017 IEEE International Conference on Computer Vision
  Workshops (ICCVW)}, 2017, pp. 1648--1655.

\bibitem{goodfellow2014generative}
I.~Goodfellow, J.~Pouget-Abadie, M.~Mirza, B.~Xu, D.~Warde-Farley, S.~Ozair,
  A.~Courville, and Y.~Bengio, ``{Generative Adversarial Nets},'' in
  \emph{Advances in Neural Information Processing Systems}.\hskip 1em plus
  0.5em minus 0.4em\relax Curran Associates, Inc., 2014, pp. 2672--2680.

\bibitem{li2017generative}
Y.~{Li}, S.~{Liu}, J.~{Yang}, and M.~{Yang}, ``{Generative Face Completion},''
  in \emph{2017 IEEE Conference on Computer Vision and Pattern Recognition
  (CVPR)}, 2017, pp. 5892--5900.

\bibitem{zhao2018robust}
F.~{Zhao}, J.~{Feng}, J.~{Zhao}, W.~{Yang}, and S.~{Yan}, ``{Robust
  LSTM-Autoencoders for Face De-Occlusion in the Wild},'' \emph{IEEE
  Transactions on Image Processing}, vol.~27, no.~2, pp. 778--790, 2018.

\bibitem{hochreiter1997long}
S.~Hochreiter and J.~Schmidhuber, ``{Long Short-Term Memory},'' \emph{Neural
  Computation}, vol.~9, no.~8, pp. 1735--1780, 1997.

\bibitem{cheng2015robust}
L.~Cheng, J.~Wang, Y.~Gong, and Q.~Hou, ``{Robust Deep Auto-encoder for
  Occluded Face Recognition},'' in \emph{23rd ACM international conference on
  Multimedia}, 2015, pp. 1099--1102.

\bibitem{galdran2015automatic}
A.~Galdran, D.~Pardo, A.~Pic{\'o}n, and A.~Alvarez-Gila, ``{Automatic
  Red-Channel Underwater Image Restoration},'' \emph{Journal of Visual
  Communication and Image Representation}, vol.~26, pp. 132--145, 2015.

\bibitem{hassner2015effective}
T.~Hassner, S.~Harel, E.~Paz, and R.~Enbar, ``{Effective Face Frontalization in
  Unconstrained Images},'' in \emph{2015 IEEE Conference on Computer Vision and
  Pattern Recognition (CVPR)}, 2015.

\bibitem{uwcorrect}
F.~Weinhaus, ``{Fred's ImageMagick Scripts: UWCORRECT},''
  \url{http://www.fmwconcepts.com/imagemagick/uwcorrect/index.php}, 2018,
  accessed: 10-28-2020.

\bibitem{fisheye}
G.~Mor, ``{iFish},'' \url{https://github.com/Gil-Mor/iFish}, 2020, accessed:
  10-28-2020.

\bibitem{park2020contrastive}
T.~Park, A.~A. Efros, R.~Zhang, and J.-Y. Zhu, ``{Contrastive Learning for
  Unpaired Image-to-Image Translation},'' \emph{arXiv preprint
  arXiv:2007.15651}, 2020.

\bibitem{schwalbe2005geometric}
E.~Schwalbe, ``{Geometric modeling and calibration of fisheye lens camera
  systems},'' in \emph{Panoramic Photogrammetry Workshop}, vol.~36, no. Part
  5/W8, 2005.

\bibitem{isola2017image}
P.~{Isola}, J.~{Zhu}, T.~{Zhou}, and A.~A. {Efros}, ``{Image-to-Image
  Translation with Conditional Adversarial Networks},'' in \emph{2017 IEEE
  Conference on Computer Vision and Pattern Recognition (CVPR)}, 2017, pp.
  5967--5976.

\bibitem{zhu2017unpaired}
J.~{Zhu}, T.~{Park}, P.~{Isola}, and A.~A. {Efros}, ``{Unpaired Image-to-Image
  Translation Using Cycle-Consistent Adversarial Networks},'' in \emph{2017
  IEEE International Conference on Computer Vision (ICCV)}, 2017, pp.
  2242--2251.

\bibitem{kim2017learning}
T.~Kim, M.~Cha, H.~Kim, J.~K. Lee, and J.~Kim, ``{Learning to Discover
  Cross-Domain Relations with Generative Adversarial Networks},'' in \emph{The
  34th International Conference on Machine Learning (ICML 2017)}, 2017, pp.
  1857--1865.

\bibitem{choi2020stargan}
Y.~{Choi}, Y.~{Uh}, J.~{Yoo}, and J.~W. {Ha}, ``{StarGAN v2: Diverse Image
  Synthesis for Multiple Domains},'' in \emph{2020 IEEE/CVF Conference on
  Computer Vision and Pattern Recognition (CVPR)}, 2020, pp. 8185--8194.

\bibitem{deng2019retinaface}
J.~{Deng}, J.~{Guo}, E.~{Ververas}, I.~{Kotsia}, and S.~{Zafeiriou},
  ``{RetinaFace: Single-Shot Multi-Level Face Localisation in the Wild},'' in
  \emph{2020 IEEE/CVF Conference on Computer Vision and Pattern Recognition
  (CVPR)}, 2020, pp. 5202--5211.

\bibitem{li19dual}
J.~{Li}, Y.~{Wang}, C.~{Wang}, Y.~{Tai}, J.~{Qian}, J.~{Yang}, C.~{Wang},
  J.~{Li}, and F.~{Huang}, ``{DSFD: Dual Shot Face Detector},'' in \emph{2019
  IEEE/CVF Conference on Computer Vision and Pattern Recognition (CVPR)}, 2019,
  pp. 5055--5064.

\bibitem{zhang16joint}
K.~{Zhang}, Z.~{Zhang}, Z.~{Li}, and Y.~{Qiao}, ``{Joint Face Detection and
  Alignment Using Multitask Cascaded Convolutional Networks},'' \emph{IEEE
  Signal Processing Letters}, vol.~23, no.~10, pp. 1499--1503, 2016.

\bibitem{zhang2019improved}
S.~Zhang, R.~Zhu, X.~Wang, H.~Shi, T.~Fu, S.~Wang, T.~Mei, and S.~Z. Li,
  ``{Improved Selective Refinement Network for Face Detection},'' \emph{arXiv
  preprint arXiv:1901.06651}, 2019.

\bibitem{abadi2016tensorflow}
M.~Abadi, P.~Barham, J.~Chen, Z.~Chen, A.~Davis, J.~Dean, M.~Devin,
  S.~Ghemawat, G.~Irving, M.~Isard, \emph{et~al.}, ``{TensorFlow: A System for
  Large-Scale Machine Learning},'' in \emph{12th USENIX symposium on operating
  systems design and implementation (OSDI 16)}, 2016, pp. 265--283.

\bibitem{paszke2019pytorch}
A.~Paszke, S.~Gross, F.~Massa, A.~Lerer, J.~Bradbury, G.~Chanan, T.~Killeen,
  Z.~Lin, N.~Gimelshein, L.~Antiga, A.~Desmaison, A.~Kopf, E.~Yang, Z.~DeVito,
  M.~Raison, A.~Tejani, S.~Chilamkurthy, B.~Steiner, L.~Fang, J.~Bai, and
  S.~Chintala, ``{PyTorch: An Imperative Style, High-Performance Deep Learning
  Library},'' in \emph{Advances in Neural Information Processing
  Systems}.\hskip 1em plus 0.5em minus 0.4em\relax Curran Associates, Inc.,
  2019, pp. 8024--8035.

\bibitem{sengupta16frontal}
S.~{Sengupta}, J.~{Chen}, C.~{Castillo}, V.~M. {Patel}, R.~{Chellappa}, and
  D.~W. {Jacobs}, ``{Frontal to Profile Face Verification in the Wild},'' in
  \emph{2016 IEEE Winter Conference on Applications of Computer Vision (WACV)},
  2016, pp. 1--9.

\bibitem{faces94}
L.~Spacek, ``{Facial Images: Faces94},''
  \url{http://cmp.felk.cvut.cz/~spacelib/faces/faces94.html}, 2009, accessed:
  10-28-2020.

\bibitem{he2016deep}
K.~{He}, X.~{Zhang}, S.~{Ren}, and J.~{Sun}, ``{Deep Residual Learning for
  Image Recognition},'' in \emph{2016 IEEE Conference on Computer Vision and
  Pattern Recognition (CVPR)}, 2016, pp. 770--778.

\bibitem{yang2016wider}
S.~Yang, P.~Luo, C.~C. Loy, and X.~Tang, ``{WIDER FACE: A Face Detection
  Benchmark},'' in \emph{IEEE Conference on Computer Vision and Pattern
  Recognition (CVPR)}, 2016.

\bibitem{nguyen2010cosine}
H.~V. Nguyen and L.~Bai, ``{Cosine Similarity Metric Learning for Face
  Verification},'' in \emph{Asian Conference on Computer Vision}.\hskip 1em
  plus 0.5em minus 0.4em\relax Springer, 2010, pp. 709--720.

\bibitem{sandler2018mobilenetv2}
M.~{Sandler}, A.~{Howard}, M.~{Zhu}, A.~{Zhmoginov}, and L.~{Chen},
  ``{MobileNetV2: Inverted Residuals and Linear Bottlenecks},'' in \emph{2018
  IEEE/CVF Conference on Computer Vision and Pattern Recognition}, 2018, pp.
  4510--4520.

\bibitem{wold1987principal}
S.~Wold, K.~Esbensen, and P.~Geladi, ``{Principal component analysis},''
  \emph{Chemometrics and intelligent laboratory systems}, vol.~2, no. 1-3, pp.
  37--52, 1987.

\bibitem{masi2018deep}
I.~Masi, Y.~Wu, T.~Hassner, and P.~Natarajan, ``{Deep Face Recognition: A
  Survey},'' in \emph{2018 31st SIBGRAPI conference on graphics, patterns and
  images (SIBGRAPI)}.\hskip 1em plus 0.5em minus 0.4em\relax IEEE, 2018, pp.
  471--478.

\bibitem{huang2007labeled}
G.~B. Huang, M.~Mattar, T.~Berg, and E.~Learned-Miller, ``{Labeled Faces in the
  Wild: A Database for Studying Face Recognition in Unconstrained
  Environments},'' University of Massachusetts, Amherst, Tech. Rep. 07--49,
  2007.

\end{thebibliography}

\end{document}